\documentclass[10pt,twocolumn,letterpaper]{article}

\usepackage[pagenumbers]{cvpr} 

\usepackage{graphicx}
\usepackage{amsmath}
\usepackage{amssymb}
\usepackage{booktabs}

\usepackage[pagebackref,breaklinks,colorlinks]{hyperref}

\usepackage{xcolor}
\usepackage{algorithm}
\usepackage{algorithmic}
\usepackage{multirow}
\usepackage{pifont}

\newcommand{\cmark}{\ding{51}}
\newcommand{\xmark}{\ding{55}}

\usepackage[capitalize]{cleveref}
\crefname{section}{Sec.}{Secs.}
\Crefname{section}{Section}{Sections}
\Crefname{table}{Table}{Tables}
\crefname{table}{Tab.}{Tabs.}

\def\cvprPaperID{7} 
\def\confName{CVPR}
\def\confYear{2023}

\begin{document}

\title{M$^2$DAR: Multi-View Multi-Scale Driver Action Recognition with Vision Transformer}

\author{Yunsheng Ma$^1$, Liangqi Yuan$^1$, Amr Abdelraouf$^2$,\and
Kyungtae Han$^2$, Rohit Gupta$^2$, Zihao Li$^1$, Ziran Wang$^1$\and
$^1$Purdue University, College of Engineering
$^2$Toyota Motor North America, InfoTech Labs\and
{\tt\small \{yunsheng, liangqiy, zihao, ziran\}@purdue.edu}\and
{\tt\small\{amr.abdelraouf, kyungtae.han, rohit.gupta\}@toyota.com}
}

\maketitle
\begin{abstract}
Ensuring traffic safety and preventing accidents is a critical goal in daily driving, where the advancement of computer vision technologies can be leveraged to achieve this goal. In this paper, we present a multi-view, multi-scale framework for naturalistic driving action recognition and localization in untrimmed videos, namely M$^2$DAR, with a particular focus on detecting distracted driving behaviors. Our system features a weight-sharing, multi-scale Transformer-based action recognition network that learns robust hierarchical representations. Furthermore, we propose a new election algorithm consisting of aggregation, filtering, merging, and selection processes to refine the preliminary results from the action recognition module across multiple views. Extensive experiments conducted on the 7th AI City Challenge Track 3 dataset demonstrate the effectiveness of our approach, where we achieved an overlap score of 0.5921 on the A2 test set. Our source code is available at \url{https://github.com/PurdueDigitalTwin/M2DAR}.
\end{abstract}

\section{Introduction}
\label{sec:intro}
Distracted driving poses a serious threat to road safety, with approximately 8.6 fatalities occurring each day in the US, a figure that is on the rise according to the National Highway Traffic Safety Administration (NHTSA)~\cite{stewart_overview_2022}. The danger is further amplified by the increased reliance of drivers on automated driving systems, especially those classified as SAE Level 3\cite{sae_on-road_automated_vehicle_standards_committee_and_others_taxonomy_2018}. These systems enable drivers to disengage from steering and pedal control, but they must still remain vigilant and prepared to regain control of the vehicle. However, drivers are prone to losing awareness of their surroundings when not actively driving, and engaging in distractions can significantly impair their ability to retake control.

Computer vision (CV) is a crucial tool in detecting distracted driving on the road, but its effectiveness can be limited by factors such as inadequate or poor quality data. To address these challenges, the Track 3 of AI City Challenge 2023~\cite{milind_naphade_7th_2023} has released a comprehensive dataset and organized a competition on naturalistic driving action recognition (DAR). The dataset features recordings of driver actions in real-world scenarios, captured from multiple camera angles, including instances of drowsy or distracted driving~\cite{naphade_6th_2022}. By analyzing these rich driving data, we can gain valuable insights into driver behavior, which can help in developing more effective driver monitoring to improve road safety. The competition's objective is not only to accurately classify but also to localize action segments within an untrimmed video sequence, a problem known as temporal action localization (TAL).


To tackle the challenges associated with DAR, we present a multi-view, multi-scale framework utilizing Vision Transformers (ViT), namely M$^2$DAR. The primary contributions of this paper include:

\begin{itemize}
\item The introduction of a weight-sharing, multi-scale Transformer-based action recognition network that learns robust hierarchical representations across multiple views.
\item A novel election algorithm consisting of four crucial steps - aggregation, filtering, merging, and selection - designed to refine the preliminary findings from the action recognition network.
\item The achievement of our proposed system, which secured 5th place on the public leaderboard of the A2 test set in the AI City Challenge 2023 Track 3, highlights the effectiveness and efficacy of our approach in accurately recognizing driver distractions.
\end{itemize}

\begin{figure*}[t]
\vskip 0.2in
\begin{center}
\centerline{\includegraphics[width=\linewidth]{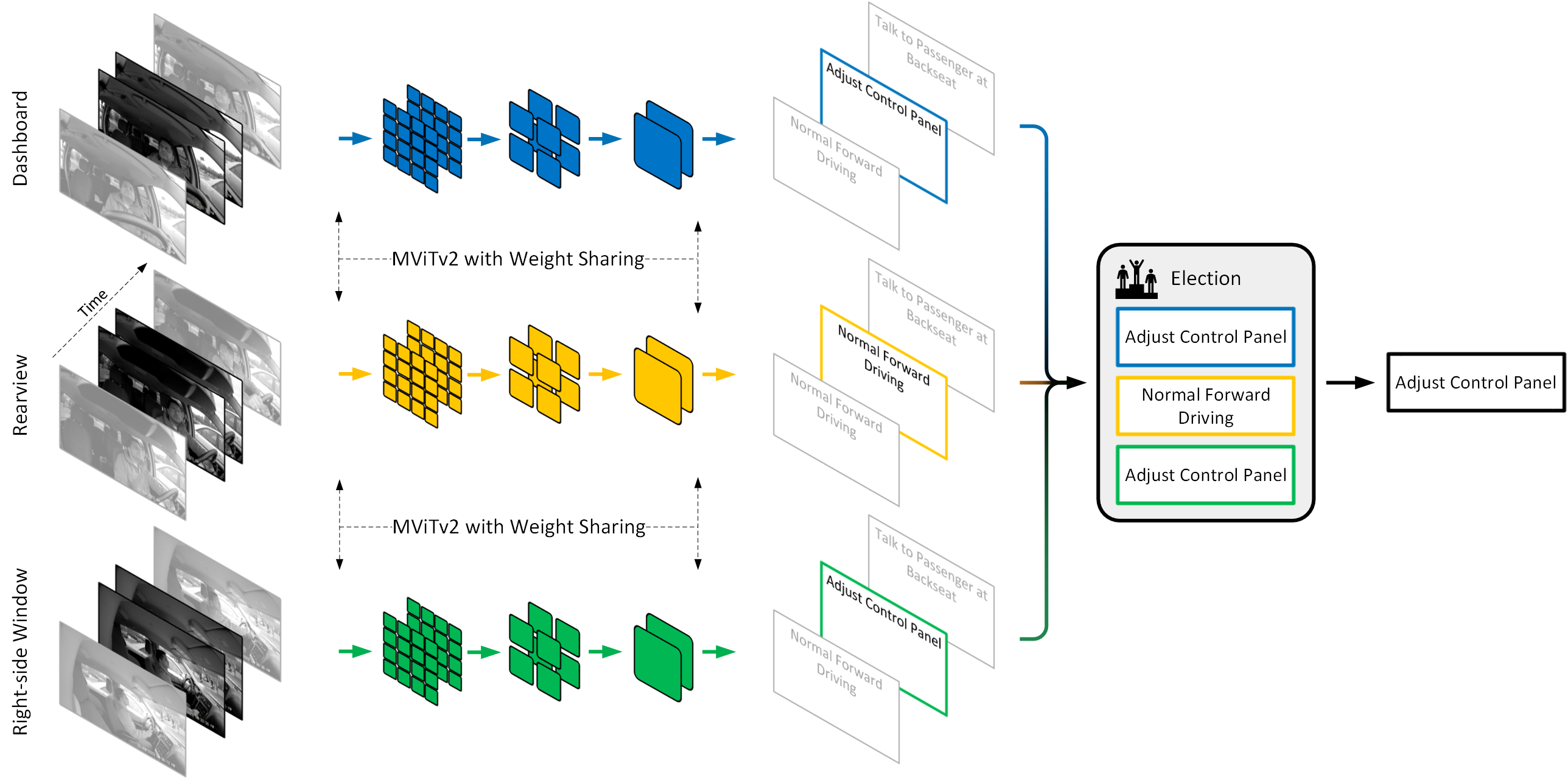}}
\caption{Schematic diagram of our M$^2$DAR system. The system consists of two stages: the action recognition stage and the election stage. In the action recognition stage, the system recognizes driver actions using a weight-sharing recognition network. In the election stage, the system refines the preliminary results obtained from the action recognition stage to generate the final action time chunks.}
\label{fig:framework}
\end{center}
\vskip -0.2in
\end{figure*}

\newcommand{\x}{\mathbf{x}}
\newcommand{\vvv}{\mathbf{v}}
\newcommand{\y}{\mathbf{y}}
\newcommand{\z}{\mathbf{z}}
\newcommand{\p}{\mathbf{p}}
\newcommand{\w}{\mathbf{w}}
\newcommand{\yhat}{\mathbf{\hat{y}}}
\newcommand{\C}{\mathcal{C}}
\newcommand{\D}{\mathcal{D}}
\newcommand{\shat}{{\hat{s}}}
\newcommand{\ehat}{{\hat{e}}}
\newcommand{\chat}{{\hat{c}}}

\section{Methodology}
In this section, we introduce our M$^2$DAR system, which offers an efficient and effective solution for accurately detecting and recognizing naturalistic driver actions. Our system consists of two stages: the DAR stage and the election stage, as illustrated in \Cref{fig:framework}.

In the \textit{DAR stage}, we use a sliding window technique with temporally overlapping frames to classify video clips of a fixed length into different action categories. This allows us to process long video sequences and identify the actions being performed within them. Our approach is designed to leverage both spatial and temporal information and effectively capture the spatiotemporal characteristics of the actions.

In the \textit{Election stage}, we refine the preliminary results obtained from the action recognition module to arrive at a final prediction. This stage is critical for improving the performance of DAR, as it allows us to consolidate the information from different camera views and select the most reliable action candidates.

\subsection{Problem Definition}
Our goal is to accurately determine the start and end times and identify the specific actions performed by a driver in each video, using input from multiple camera angles. We represent the number of camera views as $M$, and a multi-view video as $\vvv = (\x_1, ..., \x_t, ..., \x_T)$, where $\x_t$ denotes a multi-view frame. Specifically, $\x_t$ is defined as:

\begin{equation}
\x_t = \left\{\x_{t,m} \in\mathbb{R}^{C\times H_m\times W_m}\right\}_{m=1}^M,
\end{equation}
where $H_m$ and $W_m$ denote the height and width of the input image captured from view $m$, respectively, while $C$ represents the number of color channels.

Let $\y$ represent the ground-truth set of actions performed by the driver in the video, and let $\C$ be the set of predefined action categories. Each element $i$ in the ground-truth set can be expressed as $\y_i = (s_i, e_i, c_i)$, where $s_i$ indicates the starting time, $e_i$ corresponds to the ending time, and $c_i \in \C$ is the associated activity label. Let $\yhat$ be the set of $N$ predictions, where $N$ is the cardinality of $\C$\footnote{Assuming that the driver performs each of the 16 different tasks once, in random order, as stated in the challenge statement.}. We evaluate the performance of our system using the average activity overlap score.

To compute this score, we need to find a bipartite matching between the ground-truth set $\y$ and the predicted set $\yhat$, yielding a permutation of $N$ elements $\sigma \in \mathfrak{S}_{N}$ with the highest overlap score:
\begin{equation}
    \hat{\sigma}=\operatorname{argmax}_{\sigma\in\mathfrak{S}_{N}}\sum_{i=1}^N os(\y_i,\yhat_{\sigma(i)}),
\end{equation}
where $os(\y_i, \yhat_{\sigma(i)})$ is the pair-wise overlap score between the ground truth $y_i$ and a prediction with index $\sigma(i)$. A match is counted if $\y_i$ and $\yhat_{\sigma(i)}$ are of the same class ($c_i = \chat_{\sigma(i)}$) and the start time $\shat_{\sigma(i)}$ and end time $\ehat_{\sigma(i)}$ are within 10 seconds before or after the ground-truth activity's start time $s_i$ and end time $e_i$, respectively. The overlap score between each matched pair of activities is calculated as the ratio of their time intersection to their time union, as defined below:

\begin{equation}
os(\y_i, \yhat_{\sigma(i)}) = \frac{\max(\min(e_i, \ehat_{\sigma(i)}) - \max(s_i, \shat_{\sigma(i)}), 0)}{\max(e_i, \ehat_{\sigma(i)}) - \min(s_i, \shat_{\sigma(i)})}.
\label{eq:os}
\end{equation}

If a ground-truth activity has no match or a predicted activity has no ground-truth match, an overlap score of 0 is assigned. The final score is obtained by computing the average overlap score across all matched and unmatched activities.

\subsection{Driver Action Recognition Stage}
Accurate recognition of distracted driving behaviors demands a robust video classification backbone. While Transformers were initially developed for natural language processing tasks~\cite{vaswani_attention_2017}, recent progress has demonstrated their versatility beyond language tasks. For instance, ViT~\cite{dosovitskiy_image_2021} have interpreted image patches as visual words, achieving competitive performance with convolutional neural network (CNN) counterparts~\cite{karen_simonyan_very_2015,he_deep_2016}. Transformers are exceptional at modeling global information and long-range dependencies, making them suitable for analyzing video data.

Multiscale Vision Transformers (MViT) have further extended the power of ViT by introducing a pooling attention mechanism that generates a feature hierarchy with multiple stages, gradually reducing from high-resolution to low-resolution. MViT has achieved state-of-the-art performance in video tasks~\cite{fan_multiscale_2021}. To leverage these advancements, we have employed the Multiscale Vision Transformer v2 (MViTv2)~\cite{li_mvitv2_2022} as the backbone model in our system for DAR. To balance efficiency and performance, we have selected MViTv2-B (B stands for Base) as the backbone model.

The recognition module takes a fixed-length video clip as input at a time. To train the backbone model, we use a temporal data augmentation technique inspired by~\cite{li_mvitv2_2022,liang_stargazer_2022}. Specifically, we extract video clips from the original video data $\vvv$ and the corresponding variable-length annotation set $\y$, and assign them with corresponding activity labels. We create our training set by taking the union of all video clips from different videos and annotation sets as follows:

\begin{equation}
\D_\text{{train}}=\bigcup_{k=1}^{N_\text{video}}\left\{(\x_{s_i}^k,\x_{s_i+1}^k,...,\x_{e_i}^k),c_i^k\right\}_{i=1}^{N},
\end{equation}
where $\x_{s_i}^k,\x_{s_i+1}^k,...,\x_{e_i}^k$ denote the frames in the video clip and $c_i^k$ is the corresponding activity label from video $\vvv^k$ and $N_\text{{video}}$ is the number of videos available for training. During this process, we discard empty segments (video clips without any annotations) to remove noisy information and ensure high-quality training data.

During training, we pass the data from the three camera views through a weight-sharing recognition network. For each video clip in the training set, we randomly sample a $S\times\tau$ frame segment that contains $S$ frames with a temporal stride of $\tau$, which forms a training batch. The sampling is performed independently for each camera view to ensure diverse training examples. We employ standard cross-entropy loss function to optimize the network parameters.

During inference, we adopt a sliding window approach with overlapping frames to generate predictions for each test video. Specifically, we use a window size of $S\times\tau$, which is the same as the input size of the action recognition backbone model, and slide the window across the entire video with a temporal stride of $S\times\tau/4$. For each window, we feed the corresponding frames from all three camera views through the weight-sharing recognition network and obtain a probability matrix for the action categories. We then average the scores across all frame positions of the entire video to obtain a probability matrix that captures the overall temporal dynamics of the video. Finally, we pass the resulting probability matrix to the election module to generate the final action time chunks. 

\subsection{Election Stage}
\begin{figure*}[t]
    \begin{minipage}{\textwidth}
    \centering
    \small
    \begin{tabular}{cc}
    \rotatebox[origin=c]{90}{(a) Adjust control panel} & \raisebox{-0.5\height}{\includegraphics[width=0.95\linewidth]{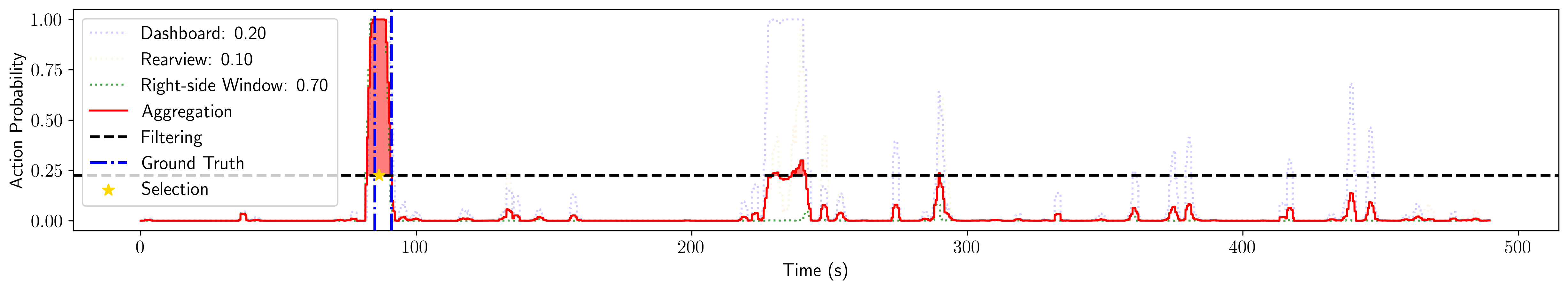}} \\
    \rotatebox[origin=c]{90}{(b) Singing} & \raisebox{-0.5\height}{\includegraphics[width=0.95\linewidth]{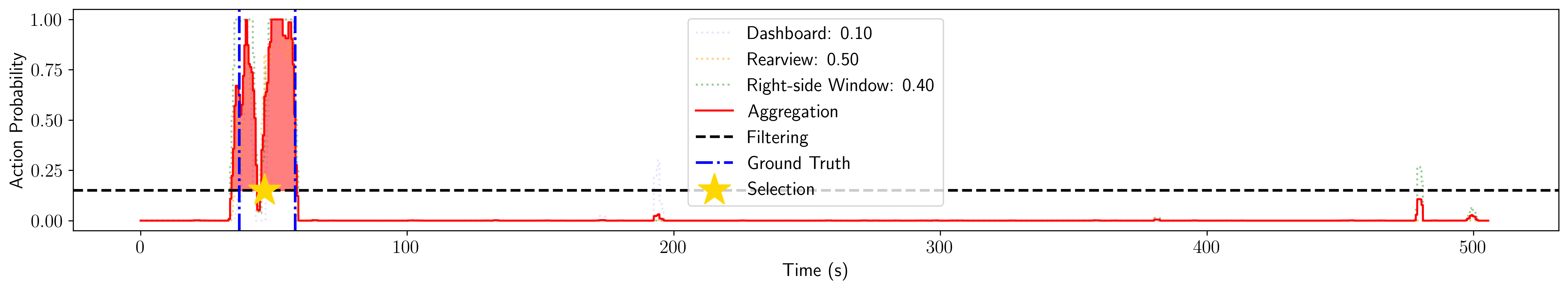}} \\
    \rotatebox[origin=c]{90}{(c) Hand on head} & \raisebox{-0.5\height}{\includegraphics[width=0.95\linewidth]{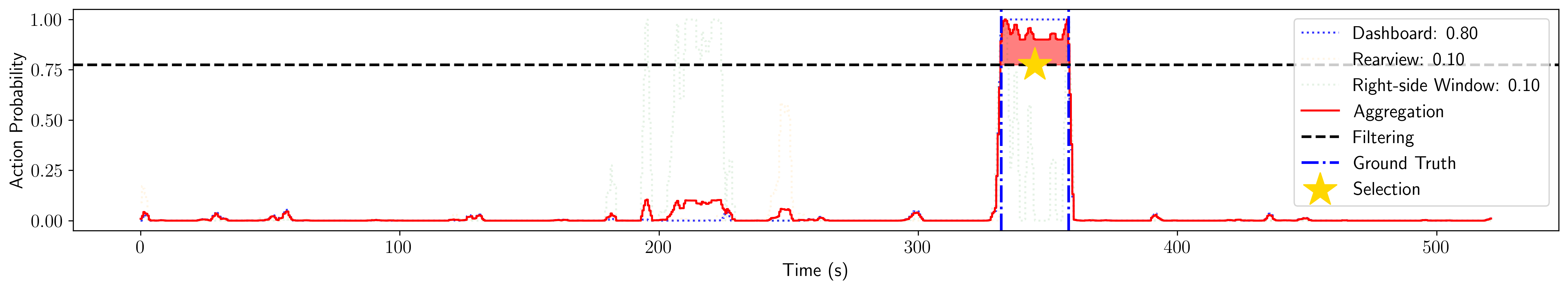}} \\
    \rotatebox[origin=c]{90}{(d) Text (right)} & \raisebox{-0.5\height}{\includegraphics[width=0.95\linewidth]{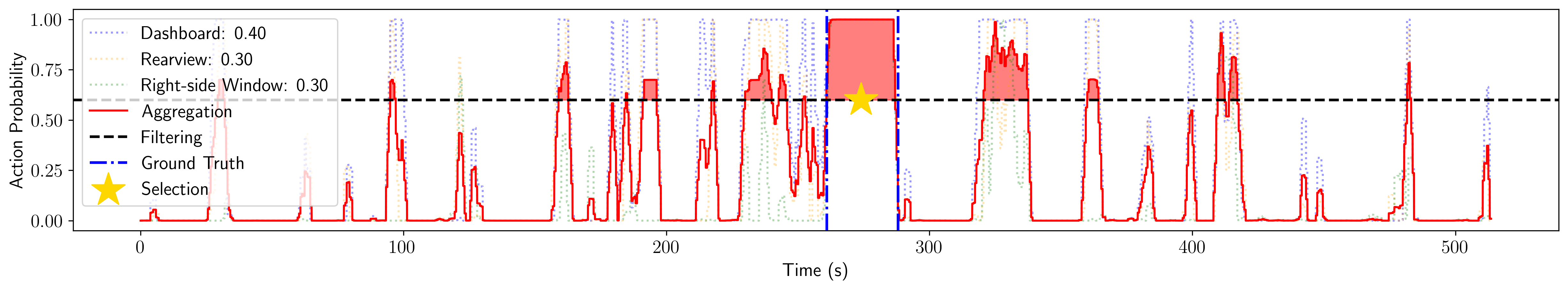}} \\
    \end{tabular}
    \end{minipage}
\caption{The figure visualizes how our proposed election algorithm consolidates preliminary findings from the action recognition module within our M$^2$DAR framework. The plot displays probability scores from the action recognition module of four action categories: (a) adjust control panel, (b) singing or dancing with music, (c) hand on head, and (d) text (right). The blue, yellow, and green dotted lines represent the outputs from the recognition module, with transparency indicating their respective weights in recognizing the action. The red line shows the probability score after the aggregation step of the election algorithm. The black dashed lines represent the probability thresholds, while the red regions between the black dashed line and the red line are the action candidates. The gold star indicates the outcome of the selection step.}
\label{fig:election}
\end{figure*}

To refine preliminary findings obtained from the action recognition module, we propose a novel algorithm called Election. The algorithm leverages a probability matrix $\p \in \mathbb{R}^{T \times |\C| \times M}$ as input, where $T$, $|\C|$, and $M$ represent the video's duration, the count of pre-defined action categories, and the number of camera views, respectively. The proposed method has four steps.

\paragraph{Aggregation (AGG).} In the first step, to capture information from various camera views, we apply a convolution operation to the input probability matrix using convolution kernels. Specifically, the operation is defined as:

\begin{equation}
\p'_{t,c}=\sum_{m=1}^{M}\omega_{c,m}\cdot \p_{t,c,m},
\end{equation}
where $\p' \in \mathbb{R}^{T \times |\C|}$ is the aggregated probability scores. To weight the information from each camera view $m$ differently for each action category $c$, we define the convolution kernels $\omega\in\mathbb{R}^{|\C| \times M}$. The kernel weight $\omega_{c,m}$ is specifically set for each action category $c$ and camera view $m$ to integrate the complementary information from different camera views while focusing on the one containing the most discriminative information. This design is based on the observation that different perspectives have different effects on various behavior recognition, and different behaviors have different characteristics under different perspectives. For example, the action \textit{talk to passenger at backseat} may be difficult to recognize from the dashboard view or the rear-view view, but very clear from the right-side window view. Therefore, the convolution operation aims to enhance the quality of the probability scores by integrating the complementary information from different camera views while weighting the views based on their relevance for each specific action category.

\paragraph{Filtering (FLTR).} In the second step, the system identifies initial action candidates by extracting continuous frames with probability scores that exceed a predefined threshold for each action category. These frames are considered as potential action segments that may contain the target actions. The threshold is set empirically based on the probability distribution on the validation data to ensure a balance between recall and precision. Frames with probabilities below this threshold are discarded as they are unlikely to represent a valid action. The resulting clips are considered as initial action candidates and are used as input for the subsequent candidate merging step.

\paragraph{Merging (MRG).} In the third step, we merge clips that have a small temporal gap (e.g., less than 0.5 seconds) between them. This design is based on the observation that some driver actions may have a significant pause during their occurrence, which can result in two separate action segments. By merging these clips, we aim to eliminate the influence of those interruption to the final action localization results. The merging process is performed by iteratively comparing the temporal distance between each pair of adjacent action candidate clips, and merging them if the distance is less than the predefined gap threshold. This process is repeated until no further merging is possible. 

\paragraph{Selection (SEL).} After the merging step, we compute the average score of all merged candidates for each action category. If there are multiple candidates for an action category, we choose the one with the highest average score as the final action candidate. The algorithm outputs $|\C|$ final action candidates, one for each action category. We round the start and end times of the final action candidates to the nearest second and output them as the system's final prediction. This process ensures that the final prediction is based on a comprehensive evaluation of all the merged candidates and their scores, resulting in a more accurate and reliable prediction of the driver's actions.

\section{Experiments}

\subsection{Dataset Description}
\begin{table}[t]
  \centering
  \begin{tabular}{lc|lc}
    \toprule
    ID & Description & ID & Description\\
    \midrule
    0 & Forward Driving & 8 & Adjust control panel\\
    1 & Drinking & 9 & Pick up from floor (D)\\
    2 & Phone Call (R) & 10 & Pick up from floor (P)\\
    3 & Phone Call (L) & 11 & Talk to pax at the right\\
    4 & Eating & 12 & Talk to pax at backseat\\
    5 & Text (R) & 13 & Yawning\\
    6 & Text (L) & 14 & Hand on head\\
    7 & Reaching behind & 15 & Singing or dancing\\
    \bottomrule
  \end{tabular}
  \caption{List of 16 driver actions defined in Track 3 of the AI City Challenge 2023: L, R, D, and P represent Left, Right, Driver, and Passenger, respectively.}
  \label{tab:actions}
\end{table}
Track 3 of the AI City Challenge 2023 \cite{milind_naphade_7th_2023} involves the analysis of synthetic naturalistic driver data captured from three different camera views positioned inside the vehicle: dashboard, right-side window, and rearview mirror, while drivers simulate driving scenarios. In addition, the drivers have three different natural driving appearances, including none, sunglasses, and hat.

The dataset consists of 34 hours of videos captured from 35 drivers performing 16 distinct driving tasks defined in \cref{tab:actions}~\cite{rahman_synthetic_2022}. Each video has a length of approximately 8 minutes, with a frame rate of 30 fps and a resolution of $1920 \times 1080$. The driver videos are divided into three subsets: A1, A2, and B. The A1 dataset is used for training and includes ground truth labels for start time, end time, and the types of distracted behaviors. The remaining two subsets, A2 and B, each containing videos from five drivers, are used for testing.

\subsection{Implementation}
We implemented the proposed system using PySlowFast~\cite{haoqi_fan_pyslowfast_2020}, an open-source codebase for video understanding. We set a consistent input size of $448\times 448$ for all three camera views, since their resolutions are identical. For training, we used all the data in the A1 subset for leaderboard submissions. We utilized a pretrained MViTv2-B model on Kinetics-700~\cite{kay_kinetics_2017}, with a frame length of $S=16$ and a sample rate of $\tau=4$. We employed the AdamW~\cite{loshchilov_decoupled_2019} optimizer with a weight decay of 0.0001, and a cosine learning rate scheduler~\cite{loshchilov_sgdr_2017}, with a base learning rate of $5\times10^{-6}$, a warm-up period of 30 epochs, and a total of 200 epochs. We conducted all training and inference processes on a NVIDIA A100 GPU.

\subsection{Main Results}
\begin{table}[t]
\centering
\begin{tabular}{cccc|c}
\toprule
\multicolumn{4}{c}{Step} \vrule & \multirow{2}{*}{Overlap Score}\\
\cmidrule{1-4}
SEL & FLTR & MRG & AGG \\
\midrule
\cmark & \xmark & \xmark & \xmark & 0.4683 \\
\cmark & \cmark & \xmark & \xmark & 0.5347 \\
\cmark & \cmark & \cmark & \xmark & 0.5565 \\
\cmark & \cmark & \cmark & \cmark & \textbf{0.5921} \\
\bottomrule
\end{tabular}
\caption{Ablation study comparing the effectiveness of individual steps in the M$^2$DAR system. SEL, FLTR, MRG, and AGG refer to the selection step, filtering step, merging step, and aggregation step, respectively. The scores shown are obtained by uploading the inference results to the evaluation server.}
\label{tab:ablation}
\end{table}

Our proposed framework achieved an overlap score (refer to \cref{eq:os}) of 0.5921 on the A2 test set. The proposed Election algorithm consists of four crucial steps: Aggregation (AGG), Filtering (FLTR), Merging (MRG), and Selection (SEL). To further validate the effectiveness of each step, we conducted an ablation study, wherein we modified one module at a time while maintaining the other modules unchanged. 

In the absence of the aggregation step, we employed the baseline method, which directly averages the probability scores from the three camera views. If the filtering step was omitted, a uniform threshold was applied to all action categories to obtain action candidates. Without the merging step, the process proceeded directly to the selection step.

The ablation study results are summarized in \cref{tab:ablation}, which presents the scores obtained after submitting the inference results to the evaluation server. As demonstrated in \cref{tab:ablation}, incorporating all four proposed modules resulted in the highest score.

\subsection{Election Visualization}
\label{sec:vis}
We present a visualization of how the proposed election algorithm consolidates the preliminary findings from the action recognition module in our M$^2$DAR framework. The visualization is shown in \cref{fig:election}, where the figures (from top to bottom) represent the probability score signals from the action recognition module on randomly selected videos from the validation set of four action categories: \textit{adjust control panel}, \textit{singing or dancing with music}, \textit{hand on head}, and \textit{text (right)}, respectively. The transparent blue, yellow, and green dot lines depict the outputs from the action recognition module, which are displayed transparently according to their weights in recognizing the particular action. The red line shows the probability score after the aggregation step of the election algorithm. The black dashed lines represent the probability thresholds. The red regions between the black dashed line and the red line are the action candidates, while the dash-dot lines represent the ground truth start and end times of the action. The gold star in the figure refers to the outcome of the selection step.

This visualization confirms the impact of different camera perspectives on recognizing different behaviors. For instance, comparing the first and third rows, we observe that the right-side window view plays a crucial role in recognizing the \textit{adjust control panel} action (first row), but can introduce significant noise in recognizing the \textit{hand on head} action (third row), which is resolved by the aggregation and filtering steps. Additionally, the second row of the \textit{singing or dancing with music} action demonstrates a pause that affects the recognition, which is addressed by the merging step of the election algorithm. Finally, the last row of the \textit{Text (Right)} action highlights how the selection step evaluates all merged candidates and their scores to ensure a comprehensive final prediction.

This visualization illustrates how our proposed algorithm effectively improves the accuracy of action recognition in multi-view videos by leveraging the complementary information from different camera views and selecting the most reliable action candidates. The integration of convolution kernels tailored for specific action categories and the merging of overlapping candidates overcomes the limitations of individual camera views and effectively captures the temporal characteristics of the actions, resulting in improved action recognition and localization performance.

\section{Related Work}
\paragraph{Driver Action Recognition.}
DAR has received significant attention from researchers due to its potential to effectively monitor driver distraction and evaluate risky driving behaviors, thereby reducing the risk and severity of traffic accidents~\cite{zhao_pand_2022,tran_effective_2022,liang_stargazer_2022,yuan2023federated,ma_vit-dd_2023, alyahya_temporal_2022,nguyen_learning_2022,vats_key_2022}. In the 6th AI City Challenge~\cite{naphade_6th_2022}, Stargazer utilized Transformer to exploit rich temporal features about human behavioral information with a simple action temporal localization framework~\cite{liang_stargazer_2022}, and PAND~\cite{zhao_pand_2022} proposed a strategy that uses a multi-branch network and a post-processing method for selecting and correcting temporal ranges. Additionally, the heterogeneity of driver behavior is also a major challenge for the generalization ability of the task~\cite{yuan2023federated}. ViT-DD proposed a semi-supervised framework to make use of driver emotion as an additional clue in recognizing driver behaviors~\cite{ma_vit-dd_2023}. Our solution for the AI City Challenge utilizes a weight-sharing approach to effectively leverage contextual information across different views, leading to better performance in multi-view scenarios.

\paragraph{Temporal Action Localization.}
TAL is a challenging task that involves accurately classifying and localizing specific activities within an untrimmed video sequence. There are two main categories of methodologies used to address this task: bi-stage methods~\cite{shou_temporal_2016,dai_temporal_2017,chao_rethinking_2018} and uni-stage methods~\cite{shou_cdc_2017,buch_end--end_2017,lin_learning_2021}. Bi-stage methods are similar to two-stage object detection approaches~\cite{ren_faster_2015,chao_rethinking_2018}, where frame or segment-level classification is performed initially, followed by a post-processing stage to consolidate the preliminary findings into a final prediction output. In contrast, uni-stage methods integrate both the temporal localization and activity categorization aspects into a single process, but the resulting complexity can lead to increased training and inference times, difficulty in scaling to handle large-scale video datasets, and meticulous fine-tuning of hyperparameters for optimal performance~\cite{gao_fine-grained_2022}. Our solution for the AI City Challenge adopts a bi-stage method by dividing the multi-view video into clips and processing them using Transformer with weight-sharing, achieving high accuracy while also reducing complexity and enabling efficient processing of large-scale video datasets.

\section{Conclusion}
In this paper, we have presented a multi-view multi-scale framework for DAR and localization in untrimmed videos, which addresses the challenges of detecting distracted driving behaviors in a naturalistic setting. The proposed M$^2$DAR framework employs a weight-sharing recognition network and an election module consisting of four steps: aggregation, filtering, merging, and selection. Our system has achieved an overlap score of 0.5921 on the A2 test set of the AI City Challenge 2023 Track 3. 
Our proposed framework has the potential to aid in the development of more effective driver monitoring systems and improve road safety.

{\small
\bibliographystyle{ieee_fullname}
\bibliography{my,liangqi-ref}
}

\end{document}